\documentclass[runningheads]{llncs}

\usepackage[pdftex]{graphicx}

\usepackage{floatrow} % For fig and table side by side
\newfloatcommand{capbtabbox}{table}[\captop][0.49\textwidth]
\usepackage{subcaption}
\usepackage{amsmath,amssymb}
\usepackage[ruled]{algorithm2e}
\usepackage{mathtools}
\usepackage{xparse, xspace} % make writing commands easier
\usepackage{microtype}
\usepackage{booktabs} % for professional tables
\usepackage{hyperref}
\usepackage[misc]{ifsym}
\usepackage{microtype}

\DeclareDocumentCommand \expectation { o m } {%
	\ensuremath{\mathop{\mathbb{E}}%
		\IfValueTF {#1} {%
			_{#1} \left[ #2 \right]%
		}{%
			\left[ #2 \right]%
		}%
	}\xspace%
}

\DeclarePairedDelimiterX{\infdivx}[2]{(}{)}{%
	#1\;\delimsize|\delimsize|\;#2%
}
\newcommand{\kld}[2]{\ensuremath{D_{KL}\infdivx{#1}{#2}}\xspace}

\DeclareMathOperator*{\argmin}{arg\,min}

\newcommand{\card}[1]{\lvert#1\rvert}

\begin{document}
\title{The KL-Divergence between a Graph Model and its Fair I-Projection as a Fairness Regularizer}
\toctitle{The KL-Divergence between a Graph Model and its Fair I-Projection as a Fairness Regularizer}

\titlerunning{KL-Divergence between a Graph Model and its Fair I-Projection}

\author{Maarten Buyl\orcidID{0000-0002-5434-2386} \Letter \and
	Tijl De Bie\orcidID{0000-0002-2692-7504}}
\tocauthor{Maarten~Buyl, Tijl~De~Bie}
\authorrunning{M. Buyl \and T. De Bie}
\institute{
	IDLab, Ghent University, Belgium\\
	\email{\{maarten.buyl, tijl.debie\}@ugent.be}}
\maketitle              % typeset the header of the contribution

\begin{abstract}
Learning and reasoning over graphs is increasingly done by means of probabilistic models, e.g. exponential random graph models, graph embedding models, and graph neural networks. When graphs are modeling relations between people, however, they will inevitably reflect biases, prejudices, and other forms of inequity and inequality. An important challenge is thus to design accurate graph modeling approaches while guaranteeing fairness according to the specific notion of fairness that the problem requires. Yet, past work on the topic remains scarce, is limited to debiasing specific graph modeling methods, and often aims to ensure fairness in an indirect manner.

We propose a generic approach applicable to most probabilistic graph modeling approaches. Specifically, we first define the class of fair graph models corresponding to a chosen set of fairness criteria. Given this, we propose a fairness regularizer defined as the KL-divergence between the graph model and its I-projection onto the set of fair models. We demonstrate that using this fairness regularizer in combination with existing graph modeling approaches efficiently trades-off fairness with accuracy, whereas the state-of-the-art models can only make this trade-off for the fairness criterion that they were specifically designed for.

\end{abstract}

%========================================================
%========================================================
\section{Introduction}
%========================================================
%========================================================

Graphs are flexible data structures, naturally suited for representing relations between people (e.g. in social networks) or between people and objects (e.g. in recommender systems). Here, links between nodes may represent any kind of relation, such as interest or similarity. It is common in real-world relational data that the corresponding graphs are often imperfect or only partially observed. For example, it may contain spurious or missing edges, or some node pairs may be explicitly marked as having unknown status. In such cases, it is often useful to correct or predict the link status between any given pair of nodes. This task is known as \textit{link prediction}: predicting the link status between any pair of nodes, given the known part of the graph and possibly any node or edge features \cite{liben2007link}.

Methods for link prediction are typically based on machine learning. A first class of methods constructs a set of features for each node-pair, such as their number of common neighbors, the Jaccard similarity between their neighborhoods, and more \cite{martinez2016survey}. Other methods are based on probabilistic models, with exponential random graph models as a notable class originating mostly from the statistics and physics communities \cite{robins2007introduction}. More recently, the machine learning community has proposed graph embedding methods \cite{hamilton2017representation}, which represent each node as a point in a vector space, from which a probabilistic model for the graph's edges can be derived (among other possible uses). Related to this, graph neural network models \cite{wu2020comprehensive} have been proposed which equally can be used to probabilistically model the presence or absence of edges in a graph \cite{zhang2018link}.

The use of such models can have genuine impact on the lives of the individuals concerned. For example, a graph of data on job seekers and job vacancies can be used to determine which career opportunities an individual will be recommended. If it is a social network, it may determine which friendships are being recommended. The existence of particular undesirable biases in such networks (e.g. people with certain demographics being recommended only certain types of jobs, or people with a certain social position only being recommended friendships with people of similar status) may result in biased link predictions that perpetuate inequity in society. Yet, graph models used for link prediction typically exploit properties of graphs that are a direct or indirect result of those existing biases. For example, many will exploit the presence of \textit{homophily}: the tendency of people to associate with similar individuals \cite{mcpherson2001birds}. However, homophily leads to segregation, which often adversely affects minority groups \cite{hofstra2017sources,karimi2018homophily}. 

The mitigation of bias in machine learning algorithms has been studied quite extensively for classification and regression models in the fairness literature, both in formalizing a range of fairness measures \cite{dwork2012fairness,hardt2016equality} and in developing methods that ensure fair classification and regression \cite{mehrabi2019survey}.
However, despite the existence of biases, such as homophily, that are specific to relational datasets, fairness has so far received limited attention in the graph modeling and link prediction literature. Current approaches focus on resolving bias issues for \emph{specific algorithms} \cite{buyl2020debayes,bose2019compositional}, or use adversarial learning to improve a \emph{specific notion of fairness} \cite{masrour2020bursting,bose2019compositional}.

\subsubsection{Contributions}
In this paper, we introduce a regularization approach to ensure fairness in link prediction that is \emph{generically} applicable across \emph{different link prediction fairness notions} and \emph{different network models}.

To that end, in Sec.~\ref{sec:fair_ipro} we first express the set of all \emph{fair probabilistic network models}. For any possibly biased network model, we can then compute the \textit{I-projection} \cite{csiszar2003information} onto this class: the distribution within the class of fair models that has the smallest KL-divergence with the biased model. In an information-theoretic sense, this I-projection can be seen as the \textit{fair} distribution that is closest to the considered biased model. We also show that for common fairness metrics, the set of fair graph models is a linear set, for which the computation of the I-projection is well-studied and easy to compute in practice.

In Sec.~\ref{sec:fair_reg}, we then propose the KL-divergence between a (possibly biased) fitted probabilistic network model and its fair I-projection as a generic \emph{fairness regularizer}, to be minimized in combination with the usual cost function for the network model. We also propose and analyze a generic algorithmic approach to efficiently solve the resulting fairness-regularized optimization problem.

Finally, our empirical results in Sec.~\ref{sec:experiments} demonstrate that our proposed fairness regularizer \emph{can be applied to a wide diversity of probabilistic network models} such that the desired fairness score is improved. In terms of that fairness criterion, our fairness modification outperforms \textit{DeBayes} and \textit{Compositional Fairness Constraints}, even on the models these baselines were specifically designed for.

%========================================================
%========================================================
\section{Related Work}\label{sec:related}
%========================================================
%========================================================
Fairness-aware machine learning is traditionally divided into three types \cite{mehrabi2019survey}: \textit{pre-processing} methods that involve transforming the dataset to remove bias \cite{calmon2017optimized}, \textit{in-processing} methods that try to modify the algorithm itself and \textit{post-processing} methods that transform the predictions of the model \cite{hardt2016equality}. Our method belongs to the in-processing category, because we directly modify the objective function with the aim of improving fairness. Here, one approach is to enforce constraints that keep the algorithm fair throughout the learning process \cite{woodworth2017learning}. 

The fairness-constrained optimization problem can also be solved using the method of Lagrange multipliers \cite{agarwal2018reductions,jiang2020identifying,cotter2019optimization,wei2020optimized}. This is related to the problem of finding the fair I-projection \cite{csiszar2003information}: the distribution from the set of fair distributions with the smallest KL-divergence to a reference distribution, e.g. an already trained (biased) model \cite{alghamdi2020model}. While we also compute the I-projection of the model onto the class of fair link predictors, we do not use it to transform the model directly. Instead, we consider the distance to that I-projection as a regularization term.

The work on applying fairness methods to the task of link prediction is limited. Methods \textit{DeBayes} \cite{buyl2020debayes}, \textit{Fairwalk} \cite{rahman2019fairwalk} and \textit{FairAdj} \cite{li2021on} all adapt specific graph embedding models to make them more fair. Other approaches, e.g. \textit{FLIP} \cite{masrour2020bursting} and \textit{Compositional Fairness Constraints} \cite{bose2019compositional}, rely on adversarial learning to remove sensitive information from node representations.

%========================================================
%========================================================
\section{Fair Information Projection}\label{sec:fair_ipro}
%========================================================
%========================================================

After discussing some notation in Sec.~\ref{sec:notation}, we characterize the set of fair graph models in Sec.~\ref{sec:fairness constraints}. In Sec.~\ref{sec:info_proj}, we will leverage this characterization to discuss the \textit{I-projection} onto the set of fair graph models, i.e. the distribution belonging to the set with the smallest KL-divergence to a reference distribution. 

%========================================================
\subsection{Notation}\label{sec:notation}
%========================================================

We denote a random unweighted and undirected graph without self-loops as $G=(V,E)$, with $V=\{1,2,\ldots,n\}$ the set of $n$ vertices and $E\subseteq {V\choose 2}$ the set of edges.
It is often convenient to represent the set of edges also by a symmetric adjacency matrix with zero diagonal $\mathbf{A}\in\{0,1\}^{n\times n}$ with element $a_{ij}$ at row $i$ and column $j$ equal to $1$ if $\{i,j\}\in E$ and $0$ otherwise.
An empirical graph over the same set of vertices will be denoted as $\hat{G}=(V,\hat{E})$ with adjacency matrix $\hat{\mathbf{A}}$ and adjacency indicator variables $\hat{a}_{ij}$. In some applications, $\hat{a}_{ij}$ may be unobserved and thus unknown for some $\{i,j\}$.

A probabilistic graph model $p$ for a given vertex set $V$ is a probability distribution over the set of possible edge sets $E$, or equivalently over the set of adjacency matrices $\mathbf{A}$, with $p(\mathbf{A})$ denoting the probability of the graph with adjacency matrix $\mathbf{A}$.
Probabilistic graph models are used for various purposes, but one important purpose is link prediction: the prediction of the existence of an edge (or not) connecting any given pair of nodes $i$ and $j$. This is particularly important when some elements from $\hat{\mathbf{A}}$ are unknown. But it is also useful when the empirical adjacency matrix is assumed to be noisy, in which case link prediction is used to reduce the noise.
Link prediction can be trivially done by making use of the marginal probability distribution $p_{ij}$, defined as $p_{ij}(x)=\sum_{\mathbf{A}:a_{ij}=x}p(\mathbf{A})$.

Note that many practically useful probabilistic graph models are dyadic independence models: they can be written as the product of the marginal distributions: $p(\mathbf{A})=\prod_{i<j}p_{ij}(a_{ij})$.
This is true for the models evaluated in our empirical results section, but the approach proposed in this paper is conceptually applicable also where this is not the case (e.g. for more complex random graph models), albeit at the cost of greater mathematical and computational complexity.

Finally, we assume vertices belong to one of a set of \emph{sensitive groups}, defined by categorical attributes with respect to which discrimination is undesirable or forbidden. These sensitive groups are denoted as $V_s$ with $s\in S$ for some finite set $S$. The sets $V_s$ with $s\in S$ form a partition of $V$. For notational convenience, we also introduce the notation $U_{st}\triangleq\{\{i,j\}|i\in V_s, j\in V_t, i\neq j\}$, the set of possible unordered pairs of distinct vertex pairs between $V_s$ and $V_t$. Thus, $|U_{ss}|={|V_s|\choose {2}}$ and $|U_{st}|=|V_s|\times |V_t|$ for $s\neq t$. Similarly, we write $U\triangleq{V\choose{2}}$ for the set of all (unordered) vertex pairs.

%========================================================
\subsection{Fairness Constraints}\label{sec:fairness constraints}
%========================================================
Here we take inspiration from prior work \cite{buyl2020debayes,laclau2021all,li2021on} on translating two classification fairness criteria to the graph setting: \textit{demographic parity} and \textit{equalized opportunity}. We then formalize a general definition for such fairness criteria.

%========================================================
\subsubsection{Demographic Parity (DP)}\label{sec:dp}

A classifier could be thought of as non-discriminatory when its expected score of an individual is the same regardless of which sensitive group they belong to. This traditional criterion of fairness is referred to as \textit{demographic} or \textit{statistical parity} (DP) \cite{dwork2012fairness}.

We generalize this to the graph setting by requiring that the expected proportion of vertex pairs belonging to any two sensitive groups $V_s$ and $V_t$ that are connected, is constant over all pairs of sensitive groups. More formally, the probabilistic graph model $p$ satisfies the DP fairness criterion iff:
\begin{align*}
\exists d \in \mathbb{R}:\forall s, t \in S:& \expectation[\mathbf{A}\sim p]{\frac{1}{|U_{st}|}\sum_{\{i,j\}\in U_{st}}a_{ij}}=d,
\end{align*}
where choices for $d$ are discussed in Sec.~\ref{sec:practical}. (Note that this criterion also ensures that the average expected vertex degree is the same for all sensitive groups.)

Thanks to linearity of the expectation operator, and with $p_{ij}$ the marginal distribution for the edge indicator variable $a_{ij}$, this can be simplified as follows:
\begin{align*}
\exists d \in \mathbb{R}: \forall s, t \in S:& \sum_{\{i,j\}\in U_{st}}\expectation[a_{ij}\sim p_{ij}]{a_{ij}}=d|U_{st}|.
\end{align*}

We thus define the set $\mathbb{P}_{\text{DP}}$ of distributions satisfying these constraints as fair with respect to DP. The DP fairness criterion is notable for diminishing the effect of homophily, since it encourages inter-group ($s \neq t$) interaction to have the same expected score as intra-group ($s = t$) interactions, thereby reducing segregation based on the nodes' sensitive traits. We note that some previous definitions \cite{laclau2021all,li2021on} enforce a weaker form of demographic parity that only requires balance between the set of all intra-group connections and the set of all inter-group connections. Quite trivially, our approach could handle this weaker form as well. However, in our experiments we maintain the stronger definition of DP fairness (defined for all pairs $\forall s, t \in S$) in order to penalize situations where one type of inter-group connections $U_{ss}$ is discriminated against in favor of a second type of inter-group connections $U_{tt} \neq U_{ss}$. 

%========================================================
\subsubsection{Equalized Opportunity (EO)}\label{sec:eo}

A drawback of the DP fairness notion is that it disregards the possibility that there are justifiable reasons for some sensitive groups to be scored higher \cite{hardt2016equality}. For example, in the social graph context one sensitive group $s$ may generally have more social interactions with others, regardless of their sensitive group $t \neq s$ \cite{buyl2020debayes}. Depending on the application, it may then be deemed fair to predict inter-group edges ($U_{st}$) from this more social group as more probable than intra-group edges between nodes in other groups ($U_{tt}$).

A fairness criterion that takes this into account is \textit{equalized opportunity} (EO) \cite{hardt2016equality}. EO requires that the true positive rate, and consequently also the false negative rate, is equal across groups. In other words, and applied to the graph context: when averaging the probability under the model of edge-connected vertex-pairs $\hat{E}$ between two sensitive groups $V_s$ and $V_t$, the result should always be the same irrespective of $s$ and $t$. More formally:
\begin{align*}
\exists d \in \mathbb{R}: \forall s, t \in S:& \expectation[\mathbf{A}\sim p]{\frac{1}{|\hat{E}\cap U_{st}|}\sum_{\{i,j\}\in \hat{E}\cap U_{st}}a_{ij}}=d,
\end{align*}
where $\hat{E}$ is the fixed empirical set of edges.

Thanks to linearity of the expectation operator, and with $p_{ij}$ the marginal distribution for the edge indicator variable $a_{ij}$, this can be simplified as follows:
\begin{align*}
\exists d \in \mathbb{R}: \forall s, t \in S:& \sum_{\{i,j\}\in \hat{E}\cap U_{st}}\expectation[a_{ij}\sim p_{ij}]{a_{ij}}=d|\hat{E}\cap U_{st}|.
\end{align*}

We thus define the set $\mathbb{P}_{\text{EO}}$ of distributions satisfying these constraints as fair with respect to EO.

%The corresponding set of distributions $\mathbb{P}_{\text{EO}}$ may be constructed using
%\begin{align*}
%\forall s \in \mathcal{S}:\quad f_s(x, y) := y \mathbf{1}_{x \in \mathcal{X}_s} \mathbf{1}_{p(y = 1 \mid x) = 1}
%\end{align*}
%where $p: \mathcal{X} \rightarrow \mathcal{Y}$ is the empirical distribution implied by $\hat{A}$. Similarly to DP, it again suffices to choose the same value $d \in [0, 1]$ for all of the constraints' constants $d_c$.

%========================================================
\subsubsection{General Sets of Fair Graph Distributions}\label{sec:general_fair_sets}

Both the DP and EO criteria are thus formalized as a constraint that is linear in the probability distribution $p$.
Using $\mathbf{1}$ to denote the indicator function, the DP and EO constraints on $p$ can both be formalized in the following form:
\begin{align}\label{eq:Fc}
F_c(p)\triangleq\sum_{\{i,j\}\in U} \expectation[a_{ij}\sim p_{ij}]{f_c(\{i,j\},a_{ij})}=d_c,
\end{align}
where for DP the functions $f_c:U\times\{0,1\}\rightarrow\mathbb{R}$ and corresponding constants $d_c$ are given by:
\begin{align*}
f_{st}(\{i,j\},x)&= x \mathbf{1}(\{i,j\}\in U_{st}),\\
d_{st}&= d|U_{st}|,
\end{align*}
for all $s,t\in S$ and for some $d\in \mathbb{R}$. Similarly, for EO:
\begin{align*}
f_{st}(\{i,j\},x)&= x \mathbf{1}(\{i,j\}\in \hat{E}\cap U_{st}),\\
d_{st}&= d|\hat{E}\cap U_{st}|.
\end{align*}
As a matter of fact, many other statistical fairness criteria, such as \textit{equalized odds}, \textit{accuracy equality} or \textit{churn equality} can formalized in this manner, with different choices for $f_c$ and $d_c$ \cite{cotter2019optimization,alghamdi2020model,agarwal2018reductions}.

Thus, although our implementation and experiments are focused on DP and EO only, we develop the theory in this paper for the general formulation of a set of fair probabilistic graph models as:\footnote{In our proposed framework, we require these constraints to be satisfied exactly in order for $p$ to be fair. However, prior work has also allowed for a percentage-wise deviation \cite{zafar2017}.} 
\begin{equation}
\mathbb{P}_\mathcal{F} := \left\{ p \in \mathbb{P} \mid \forall c \in \mathcal{C}_{\mathcal{F}}: F_c(p) = d_c \right\},
\label{eq:fair_def}
\end{equation}
with $\mathbb{P}$ the set of all possible distributions over $\mathbf{A}$, and $\mathcal{C}_{\mathcal{F}}$ a countable (and typically finite) set indexing the constraints that enforce fairness criterion $\mathcal{F}$. Importantly, $F_c$ as defined in Eq.~(\ref{eq:Fc}) is a linear function of $p$, such that I-projecting any distribution onto $\mathbb{P}_{\mathcal{F}}$ is a mathematically elegant operation. This is the subject of the following.

%To design such a constraint we introduce $F_c(q)$, the expectation of the measuring function $f_c: \mathcal{X} \times \mathcal{Y} \rightarrow \mathbb{R}$ under a probability distribution $q$:
%\begin{equation}
%F_c(q) := \expectation[x \sim \mathcal{X}]{\expectation[y \sim q(y \mid x)]{f_c(x, y)}}.
%\end{equation}\label{eq:constraints}

%The distribution $q$ can then be tested for fairness by requiring that its expected $F_c(q)$ measurements match
%\footnote{In our proposed framework, we require $F_c(q)$ to equal $d_c$ exactly in order for $q$ to be fair. However, prior work has also allowed for a percentage-wise deviation \cite{zafar2017fairness}. } 
%corresponding constants $d_c \in \mathbb{R}$. Such constraints are linear with respect to $q$. Let $\mathcal{\mathcal{C}_{\mathcal{F}}}$ be the set of these constraints that together enforce the relevant notion of what a fair probability distribution is. We can then define $\mathbb{P}_{\mathcal{F}}$, the set of distributions that match our fairness notion:
%\begin{equation}
%\mathbb{P}_{\mathcal{F}} := \left\{ q \in \mathbb{P} \mid \forall c \in \mathcal{\mathcal{C}_{\mathcal{F}}}: F_c(q) = d_c \right\}.
%\label{eq:fair_def}
%\end{equation}
%Due to the linearity of the constraints, the set $\mathbb{P}_{\mathcal{F}}$ is convex with respect to changes in $q$. 

%In what follows we review two popular notions of fairness and characterise their respective class of distributions $\mathbb{P}_{\mathcal{F}}$.

%========================================================
\subsection{Information Projection}\label{sec:info_proj}
%========================================================

%Recall the generic specification in (\ref{eq:fair_def}) of the set of fair distributions $\mathbb{P}_{\mathcal{F}}$.
We now show how to find, for any possibly unfair distribution $h$, the fair distribution $p \in \mathbb{P}_{\mathcal{F}}$ that is as close to $h$ as possible. When that closeness is computed in terms of the KL-divergence, then the desired distribution, denoted by $h_\mathcal{F}$, is known as the \textit{I-projection} \cite{csiszar1975divergence,csiszar2003information}:
\begin{align*}
h_\mathcal{F} = \argmin_{p \in \mathbb{P}_{\mathcal{F}}} \kld{p}{h},
\end{align*}
where it is assumed that $\mathbb{P}_{\mathcal{F}} \neq \emptyset$ and $\kld{p}{h} < \infty$. Since $\mathbb{P}_{\mathcal{F}}$ is linear and thus convex, the I-projection $h_\mathcal{F}$ is unique \cite{csiszar2003information}.

Finding the I-projection of model $h$ under linear constraints $\mathcal{C}_\mathcal{F}$ is a convex optimization problem \footnote{The distribution that results from the reverse KL-divergence formulation $\argmin_{p \in \mathbb{P}_{\mathcal{F}}} \kld{h}{p}$ is much less practical to compute and was therefore not further considered for this work.}. Although it is straightforward to generalize this, let us assume that $h$ is a dyadic independence model. This is justified as many contemporary probabilistic graph models (including graph embedding methods and graph neural networks) are dyadic independence models, and because it simplifies notation. Then, the I-projection of $h$ is the product distribution of the marginal distributions for the vertex pairs $\{i,j\}$, given by \cite{cover1999elements}:
\begin{equation*}
h_{\mathcal{F},ij}(x) = \frac{h_{ij}(x)}{Z_{\mathcal{F},ij}(\lambda)} \exp\left(\sum_{c \in \mathcal{C_\mathcal{F}}} \lambda_{c} f_c(\{i,j\},x)\right),
\end{equation*}%\label{eq:q_of_lag}
with 
\begin{equation*}
	Z_{\mathcal{F},ij}(\lambda) = \sum_{x\in\{0,1\}} h_{ij}(x) \exp\left(\sum_{c \in \mathcal{C_\mathcal{F}}} \lambda_{c} f_c(\{i,j\},x)\right).
\end{equation*}
the \textit{log-partition} function and with $\lambda$ denoting the vector of $\lambda_{c}$ values. Let $Z_\mathcal{F}(\lambda) = \prod_{\{i,j\}\in U} Z_{\mathcal{F},ij}(\lambda)$. The values of the $\lambda_c$ are found by maximizing:
\begin{equation}\label{eq:lambda_loss}
L_h(\lambda) = - \log Z_\mathcal{F}(\lambda) + \sum_{c \in \mathcal{C}_\mathcal{F}} \lambda_{c} d_c.
\end{equation}
%The I-projection of $h$ onto $\mathbb{P}_{\mathcal{F}}$ is thus the product distribution of $h_{\mathcal{F},ij}(\cdot;\lambda^*)$ with $\lambda^* = \argmax_\lambda L_h(\lambda)$.
This function $L_h(\lambda)$ is the Lagrange dual of the KL-divergence minimization problem with reference model $h$, and $\lambda$ is the set of Lagrange multipliers corresponding to the fairness constraints.

%========================================================
%========================================================
\section{The KL-divergence to the I-projection as a Fairness Regularizer}\label{sec:fair_reg}

We argue that the KL-divergence $\kld{h_\mathcal{F}}{h}$ between a probabilistic model $h$ and its fair I-projection $h_\mathcal{F}$ is an adequate measure of the unfairness of $h$. 

Indeed, suppose that $h_\mathcal{F}$ represents an idealized version of reality that is free from undue bias (i.e. fair). Specifically, it is the idealized version of reality that is closest to the model $h$, which, in turn, can be seen as the unfairly biased version of the reality $h_\mathcal{F}$. For example, it may be the result of discrimination and cultural social biases in historical data. Then the KL-divergence $\kld{h_\mathcal{F}}{h}$ quantifies the amount of information lost when using the biased model $h$ instead of the idealized model $h_\mathcal{F}$ \cite{burnham1998practical}. In other words, it is the information lost due to any unfairness in the model $h$, and thus, informally speaking, the amount of `unfair information' in $h$.
%Indeed, suppose an idealized version of reality is free from undue biases (i.e. fair), and $h_\mathcal{F}$ represents this idealized reality. The model $h$, on the other hand, is an unfair biased version of reality. For example, it might be the result of discrimination and cultural social biases in historical data. 

Moreover, the KL-divergence, in being a measure of information, is commensurate with commonly used loss terms in machine learning, in particular with the cross-entropy between the empirical distribution and the learned model, which is equivalent to the KL-divergence between those two up to a constant. This is the topic of the next subsection.

%========================================================
\subsection{I-Projection Regularization}
%========================================================

Let $\hat{p}$ represent the empirical distribution, i.e. $\hat{p}(\mathbf{A} = \hat{\mathbf{A}}) = 1$ and $\hat{p}(\mathbf{A} \neq \hat{\mathbf{A}}) = 0$. The common machine learning objective is then to minimize the KL-divergence $\kld{\hat{p}}{h}$, denoted by $\mathcal{L}_\mathcal{A}$, which is equivalent to maximizing the log-likelihood of $h$ under $\hat{p}$, or equivalently the cross-entropy. We propose to add the KL-divergence $\kld{h_\mathcal{F}}{h}$ as an extra loss term $\mathcal{L}_\mathcal{F}$. The overall objective function $\mathcal{L}$ to find $h$ is thus:
\begin{align*}
\mathcal{L} &= \min_h \left[\mathcal{L}_\mathcal{A} + \gamma \mathcal{L}_\mathcal{F}\right]\\
&= \min_h \left[\kld{\hat{p}}{h} + \gamma \kld{h_\mathcal{F}}{h} \right]
\end{align*}
with $\gamma$ a hyperparameter that controls the strength of the loss term. Recall that, for a parameter $\lambda$ that satisfies the fairness constraints, $\kld{h_\mathcal{F}}{h}$ is equivalent to the loss function in Eq.~(\ref{eq:lambda_loss}):
\begin{align*}
\mathcal{L} &= \min_h \left[\kld{\hat{p}}{h} + \gamma \min_{p \in \mathbb{P}_{\mathcal{F}}}\kld{p}{h}\right] \\
&= \min_h \left[\kld{\hat{p}}{h} + \gamma \max_\lambda L_h(\lambda)\right].
\end{align*}

%========================================================
\subsection{Practical Considerations}\label{sec:practical}
%========================================================

\begin{algorithm}[tb]
\KwData{possible distinct vertex pairs $U$, empirical adjacency matrix $\mathbf{\hat{A}}$, and fairness strength parameter $\gamma$} 
{\bfseries initialize} model $h$ and I-projection parameters $\lambda$\;
\For{$t=1$ {\bfseries to} $T$}{
	$\mathcal{L}_\mathcal{A} \leftarrow {-\log h\left(\mathbf{\hat{A}}\right)}$\;
	$d \leftarrow \frac{1}{\card{U}} \expectation[\mathbf{A} \sim h]{\mathbf{A}}$\;
	$\mathcal{L}_\mathcal{F} \leftarrow \max_\lambda \left[-\log Z_{h_{\mathcal{F}}}(\lambda) + \sum_{s, t \in S} \lambda_{st} d \card{U_{st}}\right]$\;
	$\mathcal{L} \leftarrow \mathcal{L}_\mathcal{A} + \gamma \mathcal{L}_\mathcal{F}$\;
	\textsc{UPDATE}$(h, \nabla_h \mathcal{L})$\;
}
\caption{Optimizing $\mathcal{L}$ with respect to link predictor $h$, in the case where DP is the fairness criterion.}
\label{alg:reg}
\end{algorithm}

So far, we did not yet specify the choice of $d$ in the DP and EO constraints. To enforce $p \in \mathbb{P}_\mathcal{DP}$, a straightforward option is to set $d$ equal to the mean of $p$. However, $d$ is then no longer constant with respect to $p$ and instead depends on changes in the $\lambda$ parameters. The gradient of the second term of the loss function $L_h(\lambda)$ in Eq.~(\ref{eq:lambda_loss}) is then more complicated. Alternatively, setting $d$ equal to the mean of the empirical distribution $\hat{p}$ forces $p$ to adopt the same mean as the empirical one, even though there is no specific reason that $h_\mathcal{F}$ or consequently $h$ should match the empirical mean. We finally chose to set $d$ equal to the mean of $h$, such that when optimizing $\lambda$, we can treat $d$ as a fixed, constant value.

Furthermore, out of several ways to optimize $\mathcal{L}$, we opted to fully optimize $\lambda$ for every parameter update of $h$. On the one hand, the $\lambda$ parameters are typically very few in number (for DP and EO, there are only $\mathcal{C}_\mathcal{F} = \card{S}^2$), making it cheap to store them. On the other hand, optimizing $\lambda$
 exactly requires the repeated evaluation of the probability under $h$ of all unordered vertex pairs $U$. With $\card{U} = \frac{n(n-1)}{2}$, this is infeasible for large $n$.
However, for $\card{S}\ll n$, using a relatively small subsample of all unordered vertex pairs will suffice in practice to obtain a good estimate for the optimal $\lambda$, dramatically enhancing scalability. Moreover, using the optimal $\lambda$ of the previous iteration's $h$ as a starting guess for the next iteration also speeds up computations in practice.
% considers all data points where $h$ is defined, i.e. the set of unordered vertex pairs $U$ with size $\card{U} = \frac{n(n-1)}{2}$. 
%Optimising with respect to all vertex pairs $U$ is therefore not scalable.
%However, the optimal $\lambda$ parameters can be approximated by only sampling a limited number of vertex pairs $h$, thereby making the method scalable to larger graphs. Using the optimal $\lambda$ of the previous iteration's $h$ as a starting guess also speeds up computations in practice.

For concreteness, the use of the proposed generic fairness regularizer to the DP fairness criterion is summarized in Alg.~\ref{alg:reg}.

%========================================================
%========================================================
\section{Experiments}\label{sec:experiments}
%========================================================
%========================================================
Our experiments were performed on three datasets, described in Sec.~\ref{sec:data}. We applied our proposed fairness regularizer on four simple, yet diverse methods explained in Sec.~\ref{sec:algos}. Though the method variants without fairness regularizer are already baselines, we additionally compared our results with state-of-the-art approaches for link prediction based on fair graph embedding in Sec.~\ref{sec:baselines}. All methods went through the same evaluation pipeline described in Sec.~\ref{sec:eval}. The results of which were discussed in Sec.~\ref{sec:results}.

%========================================================
\subsection{Datasets}\label{sec:data}
%========================================================

The methods were evaluated on three attributed graph datasets, summarized in Tab.~\ref{tab:data}. They were chosen for their diverse properties and manageable size.

\textbf{\textsc{Polblogs}}: The \textsc{Polblogs} \cite{adamic2005political} dataset was constructed from blogs discussing United States politics in 2005. In the undirected version, there is an edge between blogs if either of them had a hyperlink to the other. The sensitive attribute is the US political \textit{party} (the \textit{Republican} or \textit{Democratic} \textit{Party}) that the blog supported, either by their own admission or through manual labeling from the dataset creators. Intra-group links are heavily favored over inter-group links.

\textbf{\textsc{ML100k}}:
Movielens datasets are often used as a benchmark for recommender systems. The data contains users' movie ratings on a five-star scale. An unweighted, bipartite graph is formed by considering the users and movies as nodes and an edge between them if the user rated the movie. While the data contains several types of sensitive attributes, we opted to group the \textit{age} attribute into seven bins, delineated by the ages $[18, 25, 35, 45, 50, 56]$. There are only user-movie edges, so the domain of sensitive value of an edge is only affected by the user's sensitive value. Note that all methods were adapted such that they took the bipartitiness of the graph into account when sampling negative training edges.

\textbf{\textsc{Facebook}} \cite{mcauley2012learning}:
The \textsc{Facebook} graph consists of user nodes that are linked if they are `friends'. Each user either has \textit{gender} feature `0', `1' or neither. For the last group of users, of which there are 84, it is unclear whether their gender is unknown or non-binary. Their nodes and edges were removed from the dataset. Only 3 undirected attribute pairs thus remain in the data. In contrast to \textsc{Polblogs}, the bias effect is much weaker.

\begin{table}[tb]
\caption{Properties of the datasets. The dataset names are URLs to hosts of the datasets.}
\begin{center}
\begin{small}
\begin{sc}
\begin{tabular}{lcccc}
	\toprule
	dataset & \#nodes & \#edges & $S$ & $\card{S}$ \\
	\midrule
	\href{http://www-personal.umich.edu/\~mejn/netdata/}{Polblogs} & 1,222 & 16,714 & party & 2 \\
	\href{https://grouplens.org/datasets/movielens/100k/}{ML100k}   & 2,625 & 100,000 & age & 7 \\ 
	\href{https://snap.stanford.edu/data/egonets-Facebook.html}{Facebook} & 3,955 & 85,482 & gender & 2 \\
	\bottomrule
\end{tabular}
\end{sc}
\end{small}
\end{center}
\label{tab:data}
\end{table}

%========================================================
\subsection{Algorithms}\label{sec:algos}
%========================================================

The proposed fairness regularizer was applied to four relatively simple graph models. A \href{https://pytorch.org/}{\textit{PyTorch}} implementation was sought or implemented for each of them, such that the fairness loss can easily be added.

\textbf{\textsc{MaxEnt}}:
We will refer to the \textsc{MaxEnt} model as the maximum entropy graph model under which the expected degree of each node matches its empirical degree \cite{de2011maximum}. The solution is a simple exponential random graph model \cite{robins2007introduction}. 

\textbf{\textsc{Dot-Product}}:
Given a set of embeddings, one for every node, taking the \textsc{Dot-Product} an embedding pair is a straightforward way to perform link prediction \cite{hamilton2017representation}. In this simple model, the `decoder' for edge $(i, j)$ is the dot product operator, while the `encoder' for node $i$ just looks up its representation in a learned table of embeddings.

\textbf{\textsc{CNE}}:
A method that combines both the \textsc{MaxEnt} model and the \textsc{Dot-Product} decoder is the \textit{Conditional Network Embedding} (\textsc{CNE}) model \cite{kang2018conditional}. Instead of the Dot-Product, it `decodes' the distance between nodes $(i, j)$. Moreover, it uses the \textsc{MaxEnt} model as a prior distribution over the graph data.

\textbf{\textsc{GAE}}:
The Graph Auto-Encoder (GAE) \cite{kipf2016variational} is also a \textsc{Dot-Product} model, though it uses a Graph Convolutional Network (GCN) as its encoder. As such, it is an example of a graph neural network \cite{wu2020comprehensive}. In our implementation we used two layers for the GCN and used the identity matrix as the node feature matrix. 

%========================================================
\subsection{Fair Graph Embedding Baselines}\label{sec:baselines}
%========================================================

In part, the algorithms from Section \ref{sec:algos} were chosen such that they allow for easy comparison with two recent methods in the field of fair graph embedding.

\textbf{\textsc{CFC}}:
The Compositional Fairness Constraints (\textsc{CFC}) method \cite{bose2019compositional} aims to generate fair embeddings by learning filters that mask the sensitive attribute information. This is done through adversarial learning. When applied to link prediction, it also uses the \textsc{Dot-Product} decoder. Note that our implementation of the basic \textsc{Dot-Product} differs from the source code of \textsc{CFC}, causing differences in performance between our \textsc{Dot-Product} experiments and \textsc{CFC} with a fairness regularization strength of zero. 

\textbf{\textsc{DeBayes}}:
Finally, \textsc{DeBayes} \cite{buyl2020debayes} is an adaptation of \textsc{CNE} where the bias in the data is used as additional prior information when learning the embeddings, such that the embeddings are debiased. By using a prior without this biased information at testing time, the link prediction using these embeddings is expected to at least not be less fair than the standard \textsc{CNE}.

%========================================================
\subsection{Evaluation}\label{sec:eval}
%========================================================

Every method was run for 10 different random seeds on each dataset. Those 10 seeds each had a different train/test split, where the latter consisted of around $20\%$ of the edges in the data. The test set was extended with the same amount of non-edges. However, it was made sure that the test set did not contain nodes unknown in the train set, since the graph models in our evaluation are transductive methods. Only test set results are reported.

Hyperparameter tuning in order to improve the performance of the considered methods was minimal, as our aim is to show the effect of the fairness regularization and not the predictive quality of the methods themselves. As such, we did no hyperparameter sweep with the aim of improving AUC, and instead only deviated from default parameters when it could allow for an easier comparison between models, e.g. the dimensionality of \textsc{Dot-Product} and \textsc{CFC} embeddings. We only report results of our proposed method with a fairness regularization strength of $\gamma = 100$, because this parameter almost always caused a significant effect on the fairness measures while not diminishing predictive power too strongly. For \textsc{DeBayes} the default values were used, while for \textsc{CFC} we report the results for the regularization strength $\lambda \in \{10, 100, 1000\}$. Smaller values did not cause a noticeable effect on fairness, while larger values caused a strong degradation in terms of AUC.

Along with the link prediction AUC score, all methods were tested for their deviation from Demographic Parity (DP) and Equalized Opportunity (EO). The calculation of those measures follows \cite{buyl2020debayes}, where DP is the maximal difference between the mean predicted value of any subgroup. Similarly, the EO measure refers to the maximal difference between true positive rates of subgroups. Lower DP and EO scores therefore imply a fairer model. Note that the test set contains proportionally less negative edges than the overall dataset, possibly skewing the DP score. This effect was compensated for by proportionately increasing the contribution of negative samples when calculating DP. Furthermore, in the Appendix additional measures are reported on the diversity in the ranking of prediction scores, as well as diversity in the embeddings.

%========================================================
\subsection{Results}\label{sec:results}
%========================================================

\begin{figure*}[tbhp]
	\centering
	\begin{subfigure}[tbh]{\textwidth}
		\centering
		\includegraphics[width=\linewidth]{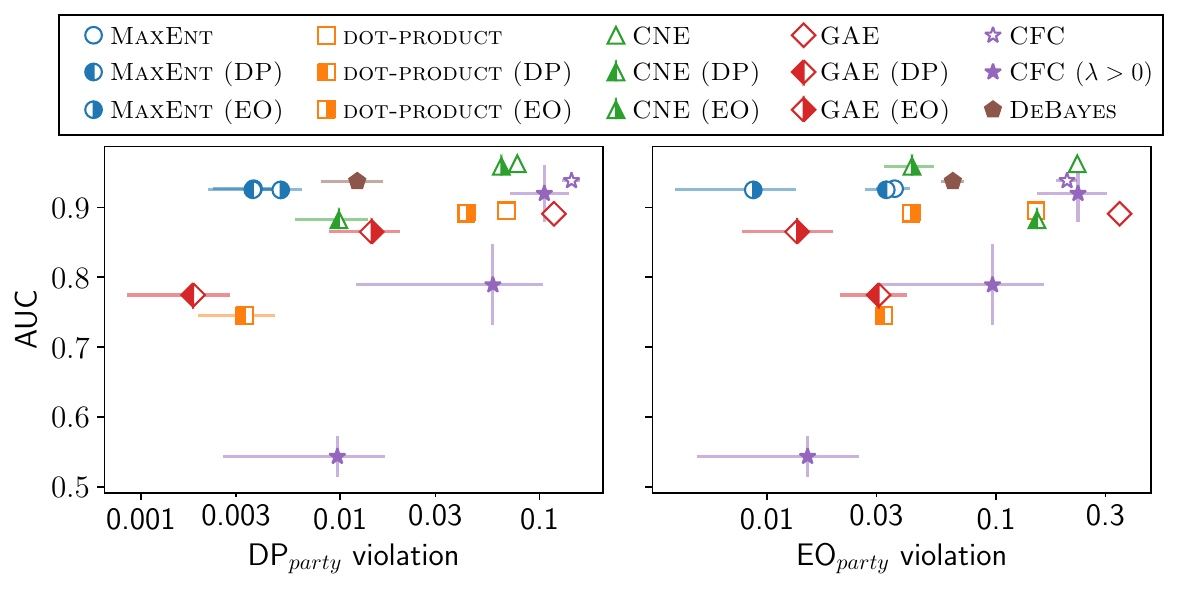}
		\caption{Results on the \textsc{Polblogs} dataset.}
		\label{fig:polblogs}
		
	\end{subfigure}%
	
	\begin{subfigure}[tbh]{\textwidth}
		\centering
		\includegraphics[width=\linewidth]{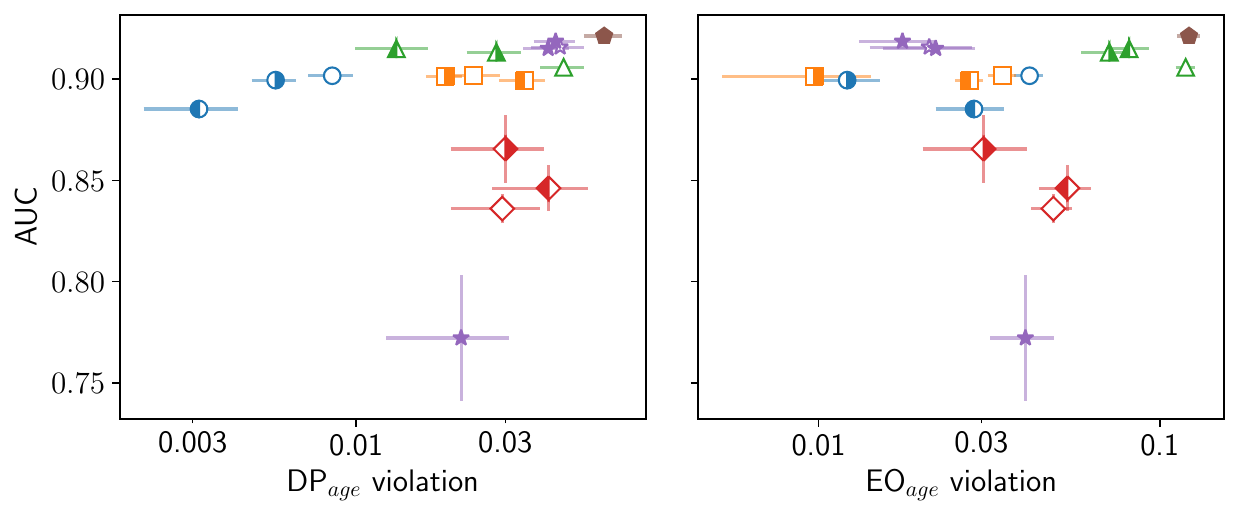}
		\caption{Results on the \textsc{ML100k} dataset.}
		\label{fig:ml100k}
	\end{subfigure}%
	
	\begin{subfigure}[tbh]{\textwidth}
		\centering
		\includegraphics[width=\linewidth]{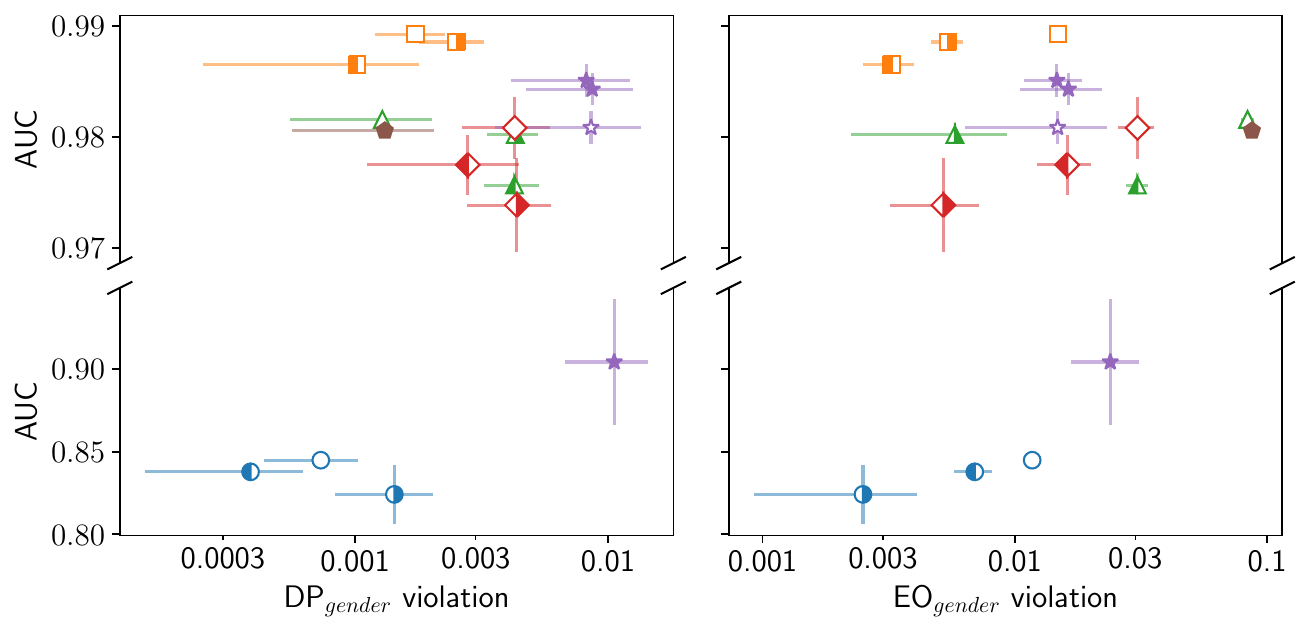}
		\caption{Results on the \textsc{Facebook} dataset.}
		\label{fig:facebook}
	\end{subfigure}%
	%\vskip 0.2in
	\caption{Markers display the mean over ten identical experiment runs with different random seeds. Error bars horizontally and vertically show the standard deviation. Completely empty markers refer to methods without any fairness modification. Methods with a fairness regularizer that enforces the DP or EO fairness criterion are left-filled or right-filled respectively. On the x-axis, unfairness is measured, so \textbf{more left is better}. On the y-axis, AUC is measured, so \textbf{higher is better}.}
	\label{fig:results}
\end{figure*}

The test set results\footnote{A table with the results in text format is provided in the Appendix.} are reported in Fig.~\ref{fig:results}. We reiterate that our intention is not to find the specific link prediction method with the best trade-off in terms of AUC and fairness. Rather, we want to verify that our proposed regularizer can be applied to a variety of methods and fairness criteria, with an efficient AUC-fairness trade-off for the considered criterion. 

\textbf{Fairness Quality}:
In many cases and across all four methods, it can indeed be observed that the use of our proposed fairness regularizer significantly reduces the link prediction bias, according to the employed fairness criteria. This is in contrast to the baselines \textsc{DeBayes} and \textsc{CFC}. The former did not improve fairness scores over \textsc{CNE}, while the latter could only become more fair at a significant cost to AUC. 

There are a few exceptions where our method does not reduce unfairness according to the fairness criterion. First, there are some cases where an already low DP score for the base method can not be improved further by adding the DP regularizer. This happens for \textsc{MaxEnt} in Fig.~\ref{fig:polblogs}, \textsc{GAE} and \textsc{Dot-Product} in Fig.~\ref{fig:ml100k} and for \textsc{CNE} in Fig.~\ref{fig:facebook}. A second kind of exception is where the method with the DP regularizer is less EO-unfair than with the EO regularizer. It occurs for the \textsc{Dot-Product (EO)} variant in Fig.~\ref{fig:polblogs} and \ref{fig:facebook}, possibly because the former had a larger reduction in predictive power overall. In both these cases, \textsc{Dot-Product (EO)} still significantly reduces EO compared to the \textsc{Dot-Product} model without fairness regularizer.

\textbf{Predictive Quality}:
Moreover, the decrease in AUC is fairly minimal with our fairness regularizer, especially compared to an adversarial approach like \textsc{CFC}. While the addition of the EO regularizer has no noticeable effect on the AUC, the DP variant does cause strong reduction on some models in Fig.~\ref{fig:polblogs}. This is to be expected, because enforcing DP can cause a significant loss in predictive power if the subgroups in the underlying data have different base rates \cite{hardt2016equality}. For a network like \textsc{Polblogs}, which strongly favors intra-group connections, encouraging the inter-group connections therefore results in AUC loss.

\begin{figure}
\CenterFloatBoxes
\begin{floatrow}
\capbtabbox{

\resizebox{0.5\textwidth}{!}{%
\begin{tabular}{l|ccc}
	\toprule
	Dataset              &   \textsc{Polblogs} &   \textsc{ML100k} &   \textsc{Facebook} \\
	\midrule
\textsc{MaxEnt}              &         14 &       68 &        158 \\
~~~with \textsc{MA}         &        707 &     3050 &       1924 \\
~~~with \textsc{EO}        &        170 &      773 &       1191 \\
\textsc{Dot-Product}         &         60 &       62 &        200 \\
~~~with \textsc{DP}    &        349 &      456 &       1169 \\
~~~with \textsc{EO}     &        135 &      239 &        531 \\
\textsc{CNE}                 &        105 &      307 &        349 \\
~~~with \textsc{DP}            &        574 &     1417 &       2065 \\
~~~with \textsc{EO}             &        286 &      843 &        865 \\
\textsc{CNE}                 &         28 &       26 &        101 \\
~~~with \textsc{DP}            &        278 &      437 &       1072 \\
~~~with \textsc{EO}         &         92 &      255 &        388 \\
\textsc{CFC}                 &        280 &      843 &       1601 \\
\textsc{CFC} $(\lambda > 0)$ &        242 &     2623 &       3494 \\
\textsc{DeBayes}             &         98 &      305 &        343 \\
\bottomrule
\end{tabular}}
%\label{tab:runtime}
}
{
	\caption{Median runtimes (s) measured by Python's \texttt{time.perf\_counter}.}\label{tab:runtime}
}

\ffigbox{
	\centering
	\includegraphics[width=0.49\textwidth]{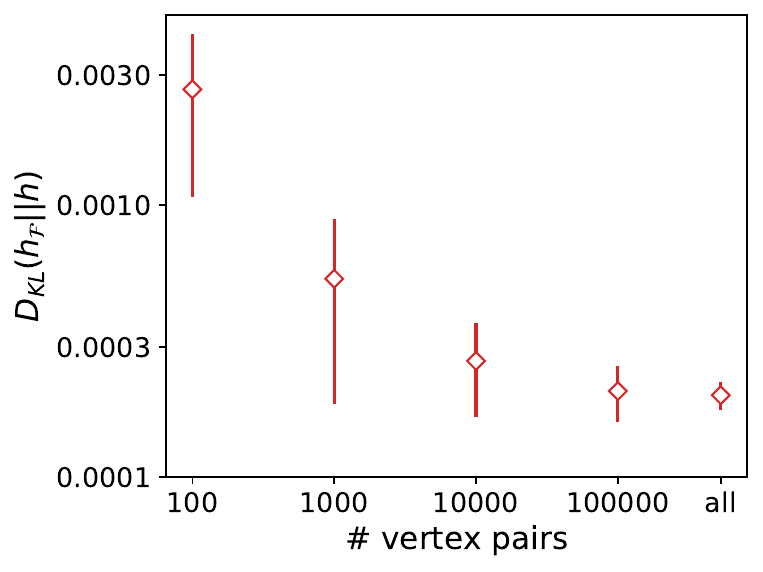}
}
{
	\caption{The KL-divergence in the experiment of Fig.~\ref{fig:facebook} between \textsc{GAE} and its fair I-projection, trained using samples from the set of all considered vertex pairs during training: all training edges plus 100 negative edges per vertex.}
	\label{fig:sampling}
}
\end{floatrow}
\end{figure}

\textbf{Runtimes}:
%Runtimes\footnote{All experiments were conducted using half the hyperthreads on a machine equipped with a 12 Core Intel(R) Xeon(R) Gold processor and 256GB of RAM} of each method are listed in Tab.~\ref{tab:runtime}. Our regularizer clearly causes a large increase in runtime, though several speed improvements are available to make the method scale to large graphs. For example, the optimal $\lambda$ parameters of $h_\mathcal{F}$ can be approximated by only fitting them on a subsample of the vertex pairs that $h$ is trained on. As shown in Fig.~\ref{fig:sampling}, the resulting KL-divergence, is already a good estimate when relatively small subsample sizes were used.  %(computed over \textit{all} vertex samples that are available to $h$)
Runtimes\footnote{All experiments were conducted using half the hyperthreads on a machine equipped with a 12 Core Intel(R) Xeon(R) Gold processor and 256GB of RAM} of each method are listed in Tab.~\ref{tab:runtime}. In our experiments, the addition of our regularizer causes a large increase in runtime. However, several easy speed improvements are available to make the method scale to large graphs. For example, the optimal $\lambda$ parameters of $h_\mathcal{F}$ can be approximated by only fitting them on a subsample of the vertex pairs that $h$ is trained on. As shown in Fig.~\ref{fig:sampling}, the resulting KL-divergence (computed over \textit{all} vertex samples that are available to $h$), is already a good estimate when relatively small subsample sizes were used.

%========================================================
%========================================================
\section{Conclusion}
%========================================================
%========================================================
Employing a generic way to characterize the set of fair link prediction distributions, we can compute the I-projection of any graph model onto this set. That distance, i.e. the KL-divergence between the model and its I-projection, can then be used as a principled regularizer during the training process and can be applied to a wide range of statistical fairness criteria.  We evaluated the benefit of our proposed method for two such criteria: demographic parity and equalized opportunity. 

Overall, our regularizer caused significant improvements in the desired fairness notions, at a relatively minimal cost in predictive power. In this it outperformed the baseline fairness modifications for graph embedding methods, which could not leverage its debiased embeddings to perform fair link prediction according to generic fairness criteria. In the future, more task-specific link prediction fairness criteria can be defined within our framework, taking inspiration from social graph or recommender systems literature. Moreover, our proposed regularizer can be extended beyond graph data structures.

%========================================================
%========================================================
\section*{Acknowledgments}
%========================================================
%========================================================
This research was funded by the ERC under the EU's 7th Framework and H2020 Programmes (ERC Grant Agreement no. 615517 and 963924), the Flemish Government (AI Research Program), the BOF of Ghent University (PhD scholarship BOF20/DOC/144), and the FWO (project no. G091017N, G0F9816N, 3G042220).
%\bibliographystyle{splncs04}
%\bibliography{references}

\clearpage
%===================================================
%===================================================
\section*{Appendix}
%===================================================
%===================================================

\renewcommand{\thesubsection}{\Alph{subsection}}
\subsection{Hyperparameters and Implementation}
As stated in the paper, hyperparameter tuning was minimal. In this section, we nevertheless provide, for each method, additional details on the choice of hyperparameters and implementations. The source code is also provided in the supplementary submission.

\textbf{\textsc{MaxEnt}} \\
The \textsc{MaxEnt} model is used in both the \textsc{CNE} and \textsc{DeBayes} models as a prior distribution. We therefore used the code available for \textsc{CNE} (available \href{https://bitbucket.org/ghentdatascience/cne/src/master/src/}{here}) as a starting point for developing a \textit{PyTorch} version of this model. The optimisation was done with \textit{LBFGS}, with at most 100 iterations.

\textbf{\textsc{Dot-Product}} \\
Due to the simplicity of the \textsc{Dot-Product} model, it was implemented from scratch in \textit{PyTorch}. A dimensionality of $128$ performed the best among the considered values $\{8, 16, 128\}$. The model was optimised with \textit{Adam} and at a learning rate of $0.01$ with $100$ iterations.

\textbf{\textsc{CNE}} \\
The \textsc{CNE} implementation was based on the code from \textsc{DeBayes}, since the latter is a fairness adaptation of \textsc{CNE}. Our \textit{PyTorch} modification was made similar to \textsc{Dot-Product}, except that it uses the distance between node embeddings instead of the actual dot product, and it uses \textsc{MaxEnt} as a prior distribution. As hyperparameters, we used the default values as listed for \textsc{DeBayes}: dimensionality of $8$, learning rate of $0.1$ and $s_2$ value of $16$. However, we found the method already converged with $200$ epochs, as opposed to the default $1000$.

\textbf{\textsc{GAE}} \\
For the \textsc{GAE} implementation, we closely followed the source code \href{https://github.com/zfjsail/gae-pytorch}{here}. We used similar parameters as \textsc{Dot-Product}, but found a dimensionality of $16$ achieved better validation scores.

\textbf{\textsc{CFC}} \\
We applied the \textsc{CFC} method to link prediction by using it as a dot-product decoder. As such, the hyperparameters are the same as for \textsc{Dot-Product}, though we stayed as close as possible to the original implementation available \href{https://github.com/joeybose/Flexible-Fairness-Constraints}{here}.

\textbf{\textsc{DeBayes}} \\
Our implementation of \textsc{DeBayes} is based on the publicly available version \href{https://github.com/aida-ugent/DeBayes}{here}. However, we used our own \textit{PyTorch} version of \textsc{CNE} in order to compare both methods easily. The \textsc{DeBayes} method has no other hyperparameters besides those listed for \textsc{CNE}.

\subsection{Additional Measures}\label{sec:add_meas}
In addition to the DP and EO measures, we evaluated the considered methods on two additional fairness measures.

\textbf{Rank DP (RDP)}\\
As stated in the paper, we computed the DP measure as the maximum difference between the mean link prediction probability of each subgroup combination, proposed in prior work \cite{buyl2020debayes}. A possible drawback of this approach is that an algorithm can reduce its DP score by simply making all prediction probabilities very similar. For example, a dot-product decoder may give privileged vertex pairs a link probability of $0.9$, and unprivileged pairs a score of $0.5$. Such a method would have a DP score of $0.4$. However, e.g. by rescaling the embeddings, a similar dot-product decoder may predict scores of $0.6$ and $0.5$ for privileged and unprivileged pairs respectively, which would result in a DP score of $0.1$. Both methods provide total separation in the \textit{ranks} of the scores between subgroups, so it could be argued that they are both extremely (yet equally) discriminatory.

Taking inspiration from prior work \cite{kallus2019fairness}, consider the AUC score, computed over the link prediction scores on the test set, when pairs of vertices of sensitive groups $V_s$ and $V_t$ respectively are the positive class and vertices with a different sensitive attribute combination are the negative class. Let the \textit{Rank Demographic Parity} (RDP) score be the maximum of those AUC scores, over all sensitive value pairs $(s, t) \in S$. Clearly, when the prediction scores are completely separated between groups as in the earlier example, the RDP score would equal $1$.

In the results reported in Sec.~\ref{sec:res_tab}, the RDP score heavily correlated with the DP score, suggesting that the latter is already an adequate measure of demographic parity in our experiment setup.

\textbf{Representation Bias (RB)}\\
The baseline methods \textsc{DeBayes} \cite{buyl2020debayes} and \textsc{CFC} \cite{bose2019compositional} focus on making the embeddings themselves less biased. While this was not the direct aim of our fairness regularizer, we evaluate all considered methods by also measuring the \textit{Representation Bias} (RB) \cite{buyl2020debayes}, i.e. the maximum AUC score that a logistic regression classifier achieves when it attempts to predict the sensitive attribute value of a vertex' embedding. 

We report the scores using such an evaluation measure in Sec.~\ref{sec:res_tab} and validate that \textsc{DeBayes} and \textsc{CFC} indeed succeed in debiasing their embeddings, as reported in their respective papers. The hyperparameters of the classifier are tuned on a validation set of vertices among all vertices in the graph. In some cases, e.g. for \textsc{Dot-Product (DP)}, the optimal hyperparameters are those that cause extreme regularisation, causing the RB scores to be $0.5$.

\subsection{Results in Table Format}\label{sec:res_tab}
The full experimental results, which were displayed as AUC-fairness trade-offs in the paper, are again displayed in Tab.~\ref{tab:polblogs}, \ref{tab:ml100k} and \ref{tab:facebook}. In addition to the DP and EO fairness measures, the RDP and RB scores as described in Sec.~\ref{sec:add_meas} are also listed.

\begin{table*}
\begin{center}
\caption{Mean $\pm$ standard deviation of results on the \textsc{polblogs} dataset.}
\label{tab:polblogs}
\fontsize{8}{10}
\begin{sc}
\begin{tabular}{l|c|c|c|c|c}
	\toprule
	Method                 & AUC                & DP                 & EO                 & RDP                & RB                 \\
	\midrule
	CFC                    & 0.938 $\pm$ 0.0047 & 0.145 $\pm$ 0.0151 & 0.204 $\pm$ 0.0211 & 0.664 $\pm$ 0.0154 & 0.977 $\pm$ 0.0049 \\
	CFC $(\lambda = 10)$   & 0.920 $\pm$ 0.0407 & 0.106 $\pm$ 0.0346 & 0.228 $\pm$ 0.0766 & 0.668 $\pm$ 0.0336 & 0.940 $\pm$ 0.1194 \\
	CFC $(\lambda = 100)$  & 0.789 $\pm$ 0.0578 & 0.058 $\pm$ 0.0463 & 0.096 $\pm$ 0.0662 & 0.626 $\pm$ 0.0442 & 0.859 $\pm$ 0.1188 \\
	CFC $(\lambda = 1000)$ & 0.544 $\pm$ 0.0297 & 0.010 $\pm$ 0.0071 & 0.015 $\pm$ 0.0101 & 0.520 $\pm$ 0.0135 & 0.595 $\pm$ 0.0482 \\
	CNE                    & 0.962 $\pm$ 0.0022 & 0.077 $\pm$ 0.0033 & 0.226 $\pm$ 0.0079 & 0.679 $\pm$ 0.0063 & 0.979 $\pm$ 0.0095 \\
	CNE (DP)               & 0.882 $\pm$ 0.0045 & 0.010 $\pm$ 0.0039 & 0.151 $\pm$ 0.0080 & 0.577 $\pm$ 0.0058 & 0.765 $\pm$ 0.0284 \\
	CNE (EO)               & 0.959 $\pm$ 0.0022 & 0.064 $\pm$ 0.0027 & 0.043 $\pm$ 0.0105 & 0.651 $\pm$ 0.0094 & 0.971 $\pm$ 0.0102 \\
	DeBayes                & 0.937 $\pm$ 0.0026 & 0.012 $\pm$ 0.0042 & 0.065 $\pm$ 0.0074 & 0.615 $\pm$ 0.0067 & 0.699 $\pm$ 0.0524 \\
	Dot-Product            & 0.895 $\pm$ 0.0035 & 0.068 $\pm$ 0.0021 & 0.149 $\pm$ 0.0056 & 0.738 $\pm$ 0.0070 & 0.975 $\pm$ 0.0062 \\
	Dot-Product (DP)       & 0.745 $\pm$ 0.0046 & 0.003 $\pm$ 0.0014 & 0.032 $\pm$ 0.0016 & 0.570 $\pm$ 0.0083 & 0.500 $\pm$ 0.0000 \\
	Dot-Product (EO)       & 0.892 $\pm$ 0.0037 & 0.043 $\pm$ 0.0016 & 0.043 $\pm$ 0.0034 & 0.657 $\pm$ 0.0076 & 0.946 $\pm$ 0.0069 \\
	GAE                    & 0.891 $\pm$ 0.0031 & 0.118 $\pm$ 0.0035 & 0.346 $\pm$ 0.0309 & 0.753 $\pm$ 0.0138 & 0.983 $\pm$ 0.0061 \\
	GAE (DP)               & 0.775 $\pm$ 0.0091 & 0.002 $\pm$ 0.0010 & 0.031 $\pm$ 0.0100 & 0.583 $\pm$ 0.0143 & 0.825 $\pm$ 0.0419 \\
	GAE (EO)               & 0.865 $\pm$ 0.0135 & 0.014 $\pm$ 0.0056 & 0.014 $\pm$ 0.0058 & 0.648 $\pm$ 0.0194 & 0.972 $\pm$ 0.0061 \\
	MaxEnt                 & 0.927 $\pm$ 0.0026 & 0.004 $\pm$ 0.0014 & 0.036 $\pm$ 0.0061 & 0.617 $\pm$ 0.0070 & / \\
	MaxEnt (DP)            & 0.925 $\pm$ 0.0028 & 0.004 $\pm$ 0.0015 & 0.033 $\pm$ 0.0065 & 0.617 $\pm$ 0.0084 & / \\
	MaxEnt (EO)            & 0.925 $\pm$ 0.0038 & 0.005 $\pm$ 0.0014 & 0.009 $\pm$ 0.0047 & 0.604 $\pm$ 0.0042 & / \\
	\bottomrule
\end{tabular}
\end{sc}
\end{center}
\end{table*}

\begin{table*}
	\begin{center}
		\caption{Mean $\pm$ standard deviation of results on the \textsc{ml100k} dataset.}
		\label{tab:ml100k}
		\fontsize{8}{10}
		\begin{sc}
			\begin{tabular}{l|c|c|c|c|c}
					\toprule
					Method                 & AUC                & DP                 & EO                 & RDP                & RB                 \\
					\midrule
					CFC                    & 0.916 $\pm$ 0.0021 & 0.045 $\pm$ 0.0086 & 0.021 $\pm$ 0.0070 & 0.531 $\pm$ 0.0040 & 0.631 $\pm$ 0.0191 \\
					CFC $(\lambda = 10)$   & 0.919 $\pm$ 0.0021 & 0.043 $\pm$ 0.0065 & 0.018 $\pm$ 0.0045 & 0.530 $\pm$ 0.0030 & 0.662 $\pm$ 0.0127 \\
					CFC $(\lambda = 100)$  & 0.915 $\pm$ 0.0023 & 0.041 $\pm$ 0.0069 & 0.022 $\pm$ 0.0066 & 0.530 $\pm$ 0.0036 & 0.681 $\pm$ 0.0158 \\
					CFC $(\lambda = 1000)$ & 0.772 $\pm$ 0.0311 & 0.022 $\pm$ 0.0092 & 0.040 $\pm$ 0.0085 & 0.519 $\pm$ 0.0035 & 0.547 $\pm$ 0.0235 \\
					CNE                    & 0.906 $\pm$ 0.0015 & 0.046 $\pm$ 0.0073 & 0.119 $\pm$ 0.0078 & 0.536 $\pm$ 0.0030 & 0.547 $\pm$ 0.0123 \\
					CNE (DP)               & 0.915 $\pm$ 0.0013 & 0.013 $\pm$ 0.0035 & 0.081 $\pm$ 0.0117 & 0.511 $\pm$ 0.0020 & 0.588 $\pm$ 0.0205 \\
					CNE (EO)               & 0.913 $\pm$ 0.0014 & 0.028 $\pm$ 0.0055 & 0.071 $\pm$ 0.0124 & 0.528 $\pm$ 0.0032 & 0.614 $\pm$ 0.0134 \\
					DeBayes                & 0.921 $\pm$ 0.0013 & 0.062 $\pm$ 0.0086 & 0.122 $\pm$ 0.0096 & 0.536 $\pm$ 0.0028 & 0.524 $\pm$ 0.0235 \\
					Dot-Product            & 0.902 $\pm$ 0.0012 & 0.024 $\pm$ 0.0051 & 0.035 $\pm$ 0.0033 & 0.528 $\pm$ 0.0029 & 0.657 $\pm$ 0.0233 \\
					Dot-Product (DP)       & 0.899 $\pm$ 0.0012 & 0.034 $\pm$ 0.0058 & 0.028 $\pm$ 0.0026 & 0.509 $\pm$ 0.0037 & 0.759 $\pm$ 0.0232 \\
					Dot-Product (EO)       & 0.901 $\pm$ 0.0012 & 0.019 $\pm$ 0.0026 & 0.010 $\pm$ 0.0045 & 0.526 $\pm$ 0.0029 & 0.697 $\pm$ 0.0275 \\
					GAE                    & 0.836 $\pm$ 0.0072 & 0.029 $\pm$ 0.0092 & 0.049 $\pm$ 0.0067 & 0.518 $\pm$ 0.0040 & 0.582 $\pm$ 0.0279 \\
					GAE (DP)               & 0.846 $\pm$ 0.0112 & 0.041 $\pm$ 0.0140 & 0.054 $\pm$ 0.0094 & 0.510 $\pm$ 0.0050 & 0.645 $\pm$ 0.0195 \\
					GAE (EO)               & 0.866 $\pm$ 0.0167 & 0.030 $\pm$ 0.0100 & 0.030 $\pm$ 0.0103 & 0.517 $\pm$ 0.0034 & 0.578 $\pm$ 0.0325 \\
					MaxEnt                 & 0.902 $\pm$ 0.0014 & 0.008 $\pm$ 0.0014 & 0.042 $\pm$ 0.0040 & 0.536 $\pm$ 0.0026 & / \\
					MaxEnt (DP)            & 0.885 $\pm$ 0.0039 & 0.003 $\pm$ 0.0011 & 0.029 $\pm$ 0.0065 & 0.515 $\pm$ 0.0055 & / \\
					MaxEnt (EO)            & 0.899 $\pm$ 0.0013 & 0.006 $\pm$ 0.0009 & 0.012 $\pm$ 0.0030 & 0.530 $\pm$ 0.0018 & / \\
					\bottomrule
				\end{tabular}
			\end{sc}
	\end{center}
\end{table*}

\begin{table*}
	\begin{center}
		\caption{Mean $\pm$ standard deviation of results on the \textsc{facebook} dataset.}
		\label{tab:facebook}
		\fontsize{8}{10}
		\begin{sc}
			\begin{tabular}{l|c|c|c|c|c}
					\toprule
					Method                 & AUC                & DP                 & EO                 & RDP                & RB                 \\
					\midrule
					CFC                    & 0.981 $\pm$ 0.0015 & 0.009 $\pm$ 0.0050 & 0.015 $\pm$ 0.0084 & 0.529 $\pm$ 0.0068 & 0.601 $\pm$ 0.0150 \\
					CFC $(\lambda = 10)$   & 0.984 $\pm$ 0.0014 & 0.009 $\pm$ 0.0039 & 0.016 $\pm$ 0.0058 & 0.529 $\pm$ 0.0040 & 0.615 $\pm$ 0.0153 \\
					CFC $(\lambda = 100)$  & 0.985 $\pm$ 0.0015 & 0.008 $\pm$ 0.0041 & 0.015 $\pm$ 0.0038 & 0.529 $\pm$ 0.0049 & 0.617 $\pm$ 0.0157 \\
					CFC $(\lambda = 1000)$ & 0.904 $\pm$ 0.0381 & 0.011 $\pm$ 0.0038 & 0.024 $\pm$ 0.0071 & 0.532 $\pm$ 0.0052 & 0.586 $\pm$ 0.0228 \\
					CNE                    & 0.982 $\pm$ 0.0006 & 0.001 $\pm$ 0.0007 & 0.084 $\pm$ 0.0050 & 0.538 $\pm$ 0.0035 & 0.601 $\pm$ 0.0189 \\
					CNE (DP)               & 0.976 $\pm$ 0.0006 & 0.004 $\pm$ 0.0010 & 0.031 $\pm$ 0.0030 & 0.512 $\pm$ 0.0027 & 0.594 $\pm$ 0.0217 \\
					CNE (EO)               & 0.980 $\pm$ 0.0006 & 0.004 $\pm$ 0.0010 & 0.006 $\pm$ 0.0035 & 0.518 $\pm$ 0.0025 & 0.607 $\pm$ 0.0192 \\
					DeBayes                & 0.981 $\pm$ 0.0004 & 0.001 $\pm$ 0.0007 & 0.087 $\pm$ 0.0042 & 0.539 $\pm$ 0.0030 & 0.592 $\pm$ 0.0163 \\
					Dot-Product            & 0.989 $\pm$ 0.0006 & 0.002 $\pm$ 0.0005 & 0.015 $\pm$ 0.0007 & 0.561 $\pm$ 0.0037 & 0.630 $\pm$ 0.0138 \\
					Dot-Product (DP)       & 0.987 $\pm$ 0.0008 & 0.001 $\pm$ 0.0008 & 0.003 $\pm$ 0.0007 & 0.515 $\pm$ 0.0032 & 0.500 $\pm$ 0.0000 \\
					Dot-Product (EO)       & 0.989 $\pm$ 0.0005 & 0.003 $\pm$ 0.0007 & 0.005 $\pm$ 0.0008 & 0.530 $\pm$ 0.0041 & 0.769 $\pm$ 0.0185 \\
					GAE                    & 0.981 $\pm$ 0.0028 & 0.004 $\pm$ 0.0016 & 0.031 $\pm$ 0.0050 & 0.545 $\pm$ 0.0037 & 0.602 $\pm$ 0.0152 \\
					GAE (DP)               & 0.977 $\pm$ 0.0027 & 0.003 $\pm$ 0.0017 & 0.016 $\pm$ 0.0039 & 0.518 $\pm$ 0.0061 & 0.626 $\pm$ 0.0193 \\
					GAE (EO)               & 0.974 $\pm$ 0.0042 & 0.004 $\pm$ 0.0016 & 0.005 $\pm$ 0.0020 & 0.535 $\pm$ 0.0026 & 0.618 $\pm$ 0.0240 \\
					MaxEnt                 & 0.845 $\pm$ 0.0014 & 0.001 $\pm$ 0.0003 & 0.012 $\pm$ 0.0007 & 0.541 $\pm$ 0.0041 & / \\
					MaxEnt (DP)            & 0.838 $\pm$ 0.0043 & 0.000 $\pm$ 0.0002 & 0.007 $\pm$ 0.0012 & 0.530 $\pm$ 0.0051 & / \\
					MaxEnt (EO)            & 0.824 $\pm$ 0.0180 & 0.001 $\pm$ 0.0006 & 0.002 $\pm$ 0.0016 & 0.519 $\pm$ 0.0070 & / \\
					\bottomrule
				\end{tabular}
			\end{sc}
	\end{center}
\end{table*}
\end{document}